\documentclass[10pt,journal,compsoc]{IEEEtran}

\ifCLASSOPTIONcompsoc
  \usepackage[nocompress]{cite}
\else
  \usepackage{cite}
\fi

\usepackage{amsmath,amssymb,graphicx,caption,subcaption,url}

\begin{document}

\title{Implicit segmentation of Kannada characters in offline handwriting recognition using hidden Markov models}

\author{Manasij Venkatesh, Vikas Majjagi, and Deepu Vijayasenan% 
\IEEEcompsocitemizethanks{\IEEEcompsocthanksitem The authors are with the Department of Electronics and Communication Engineering, National Institute of Technology Karnataka, Surathkal-575025, India.\protect\\
E-mail: manasij.ece.nitk@gmail.com, vikasm.ece.nitk@gmail.com, deepu.senan@gmail.com
}% 
} %

\IEEEtitleabstractindextext{%
\begin{abstract}
We describe a method for classification of handwritten Kannada characters using Hidden Markov Models (HMMs). Kannada script is agglutinative, where simple shapes are concatenated horizontally to form a character. This results in a large number of characters making the task of classification difficult. Character segmentation plays a significant role in reducing the number of classes. Explicit segmentation techniques suffer when overlapping shapes are present, which is common in the case of handwritten text. We use HMMs to take advantage of the agglutinative nature of Kannada script, which allows us to perform implicit segmentation of characters along with recognition. All the experiments are performed on the Chars74k dataset that consists of 657 handwritten characters collected across multiple users. Gradient-based features are extracted from individual characters and are used to train character HMMs. The use of implicit segmentation technique at the character level resulted in an improvement of around 10\%. This system also outperformed an existing system tested on the same dataset by around 16\%. Analysis based on learning curves showed that increasing the training data could result in better accuracy. Accordingly, we collected additional data and obtained an improvement of 4\% with 6 additional samples.
\end{abstract}

\begin{IEEEkeywords}
	Handwriting recognition, hidden Markov models, implicit segmentation, offline, cursive script, Kannada.
\end{IEEEkeywords}}

\maketitle

\IEEEpeerreviewmaketitle

\section{Introduction}

Even with the advent of new technologies, handwritten text continues to be a method of recording information and is also a means of communication. Tasks such as automatically interpreting postal addresses, reading bank checks and automatic processing of handwritten forms necessitates the need for a handwriting recognition system. Several open-source and commercial systems exist for the recognition of printed text \cite{google:tesseract, nuance}. However, recognition of handwritten characters is a more challenging problem as it involves variability of handwriting such as intra-writer and inter-writer differences and overlapping of characters which increases difficulty in segmentation. There is still scope for improvement in these systems.

Research in handwriting recognition has been popular in the past few decades \cite{srihari:survey, steinherz:offline}. In recent years, substantial work has been done in the field of online handwriting recognition of Indic scripts \cite{pal2004:indian, agr:kannada, agr:tamil, deepu:devanagari}. However, viable offline HWR systems have been developed for only a few languages and most Indic languages are still beyond the pale of current HWR techniques. To the authors' best knowledge, offline recognition of Kannada script has been addressed in very few works \cite{kumar:scene}. Further, no Markov model based handwriting recognition system exists for Kannada even though HMM based offline models have been shown to perform well in recognizing English \cite{hmm:eng} and Chinese handwritten text \cite{hmm:chi}. Using HMMs allows for the distinct advantage of simultaneous training and segmentation. Implicit segmentation, which leaves the task of segmentation to the recognizer, has been described by Cavalin et al \cite{cavalin:imp}. The offline system searches a word for components that match characters in its alphabet. 

\begin{figure}[t]
	\begin{center}
		\begin{tabular}{ccccc}
			\includegraphics[width=0.5in]{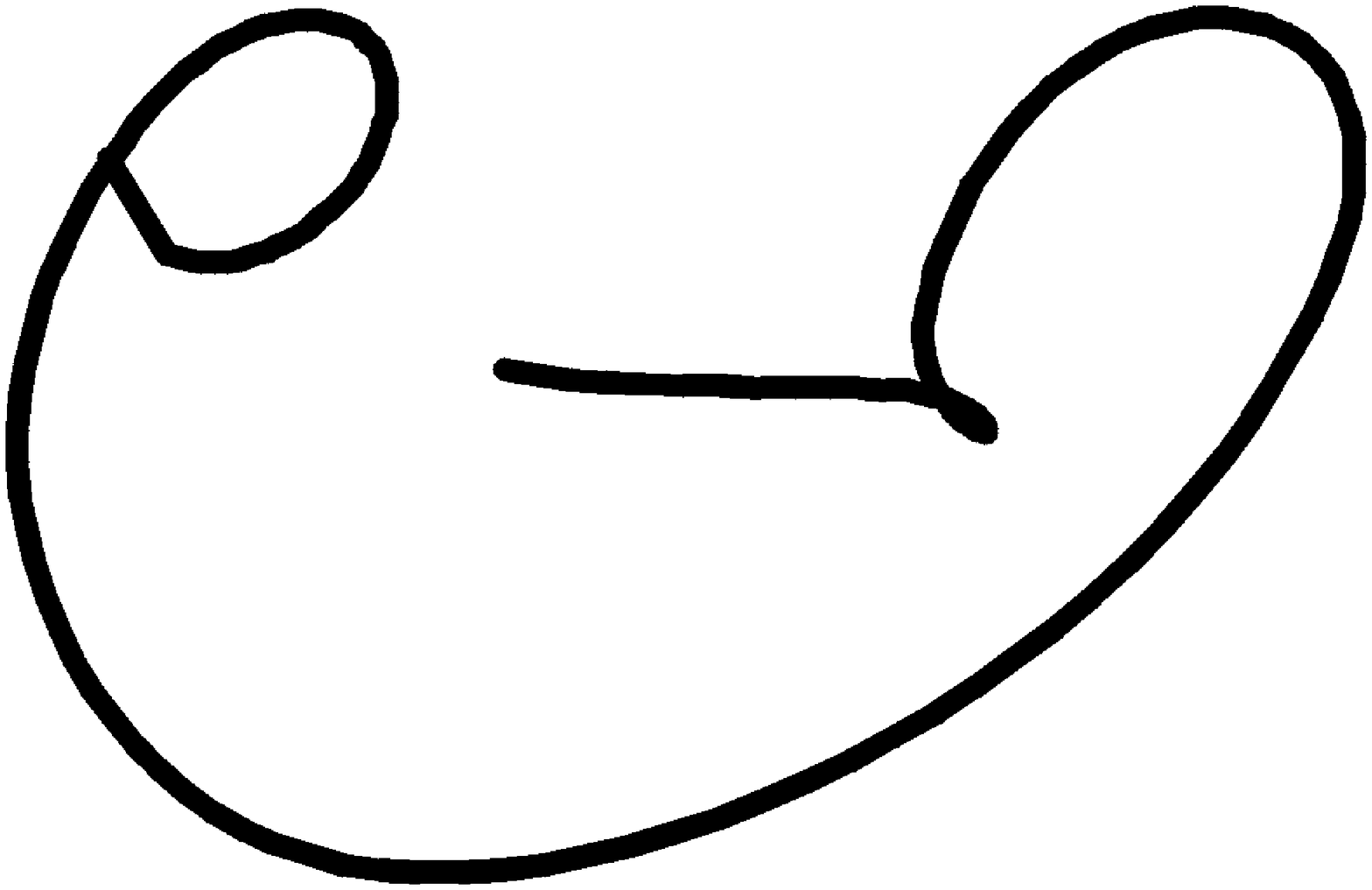} &
			\includegraphics[width=0.5in]{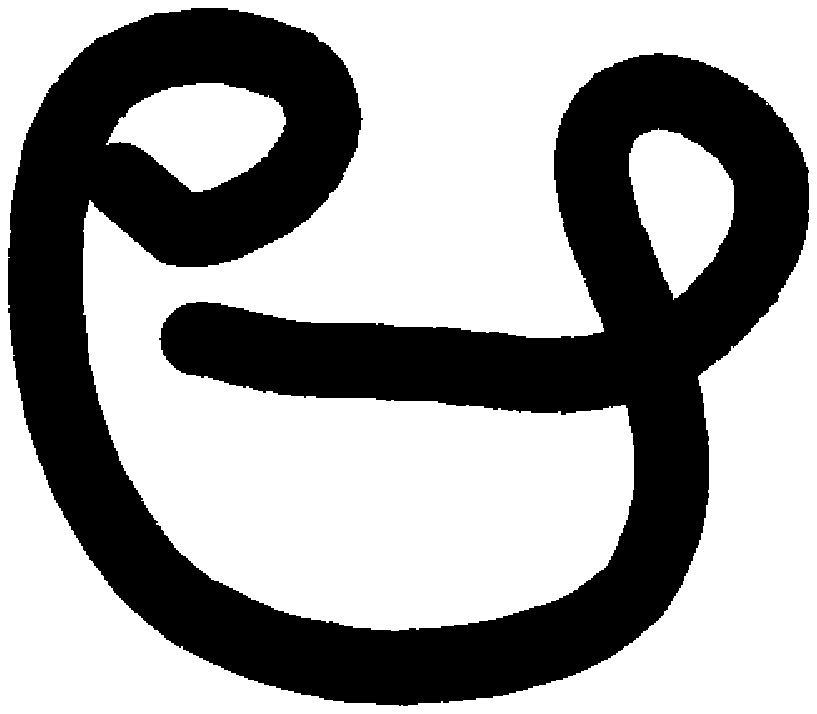} &
			\includegraphics[width=0.5in]{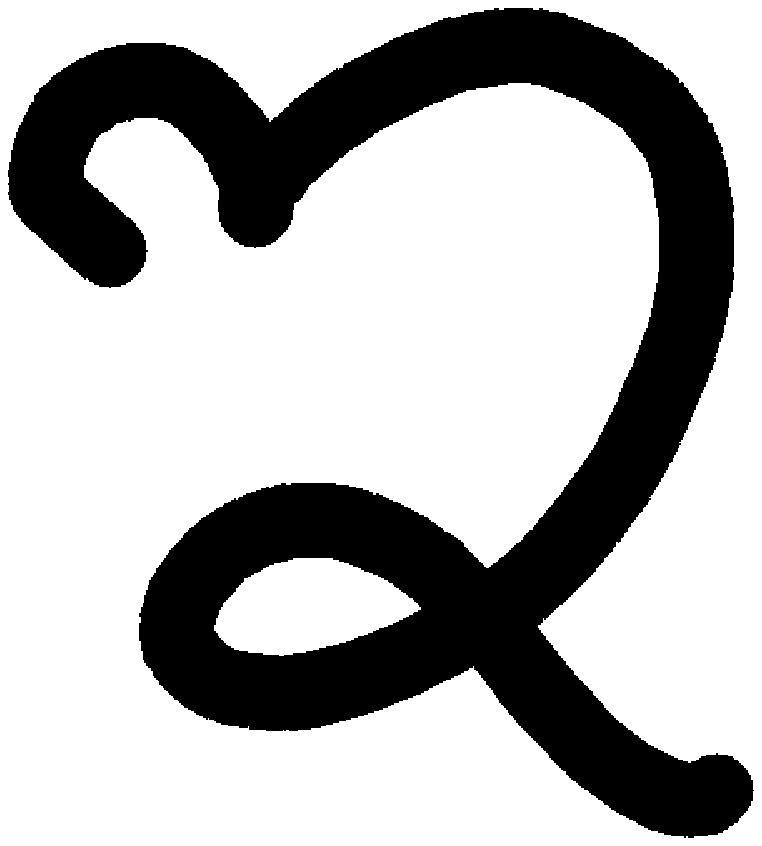} &
			\includegraphics[width=0.5in]{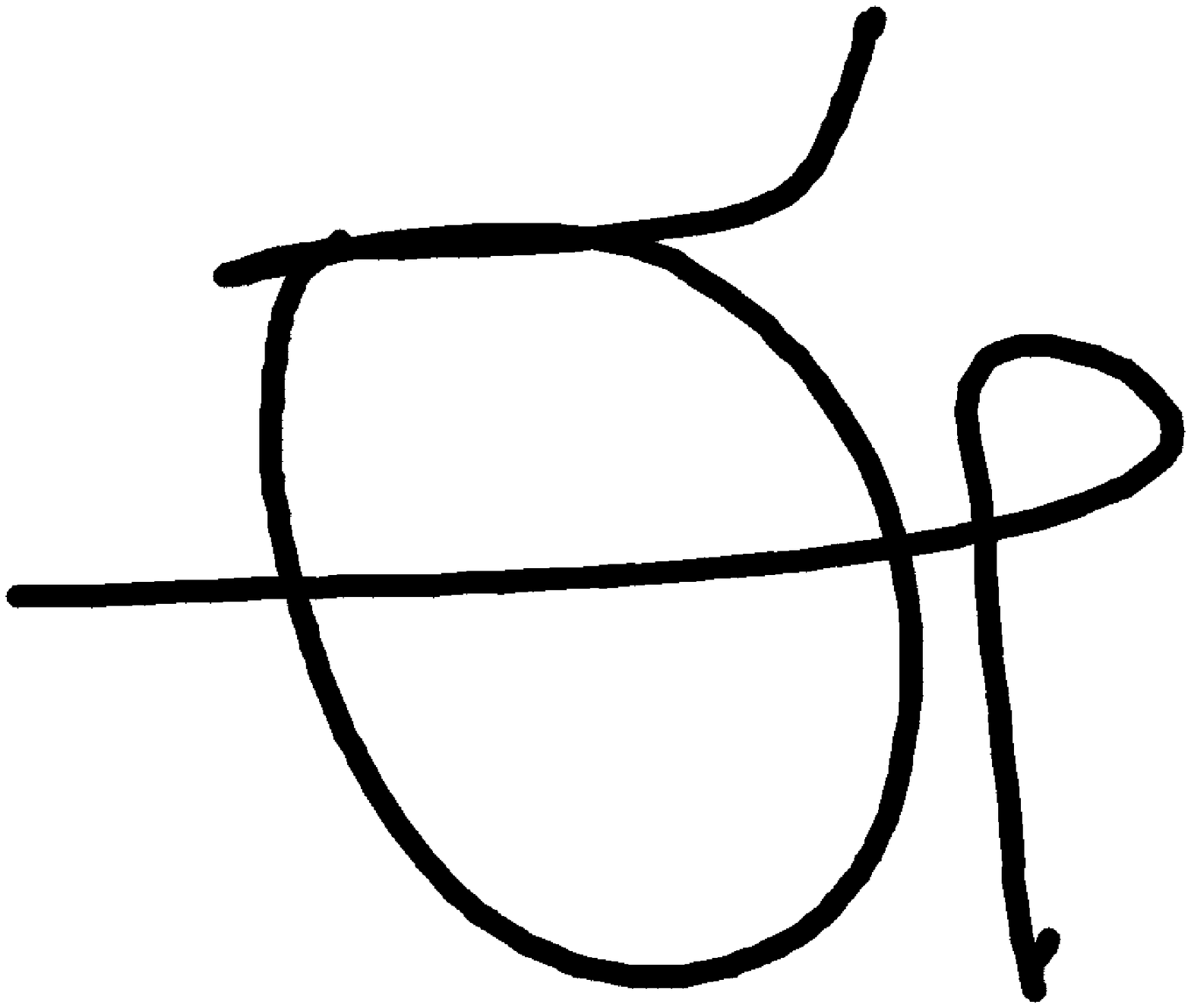} &
			\includegraphics[width=0.5in]{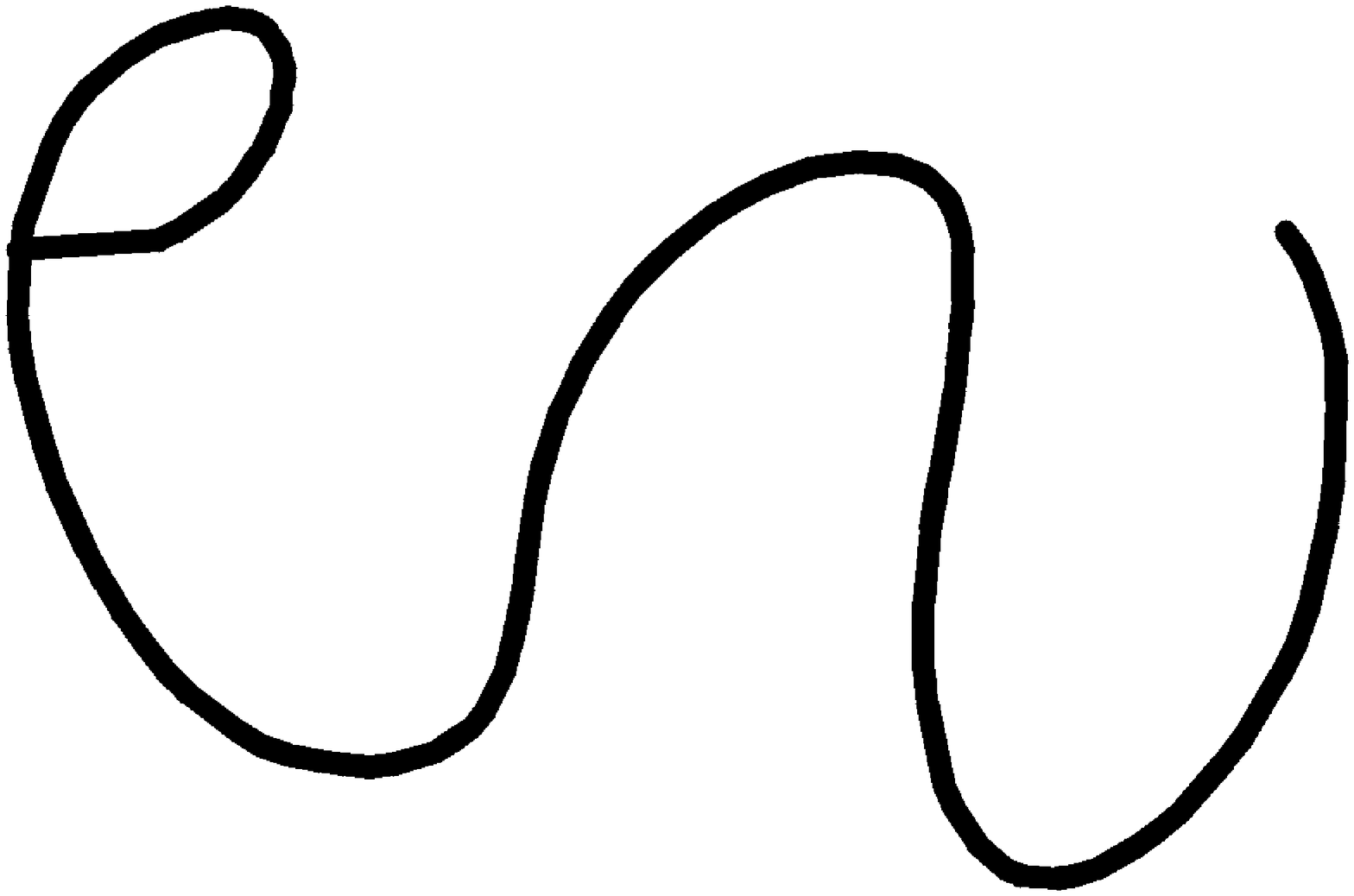} \\
			\includegraphics[width=0.5in]{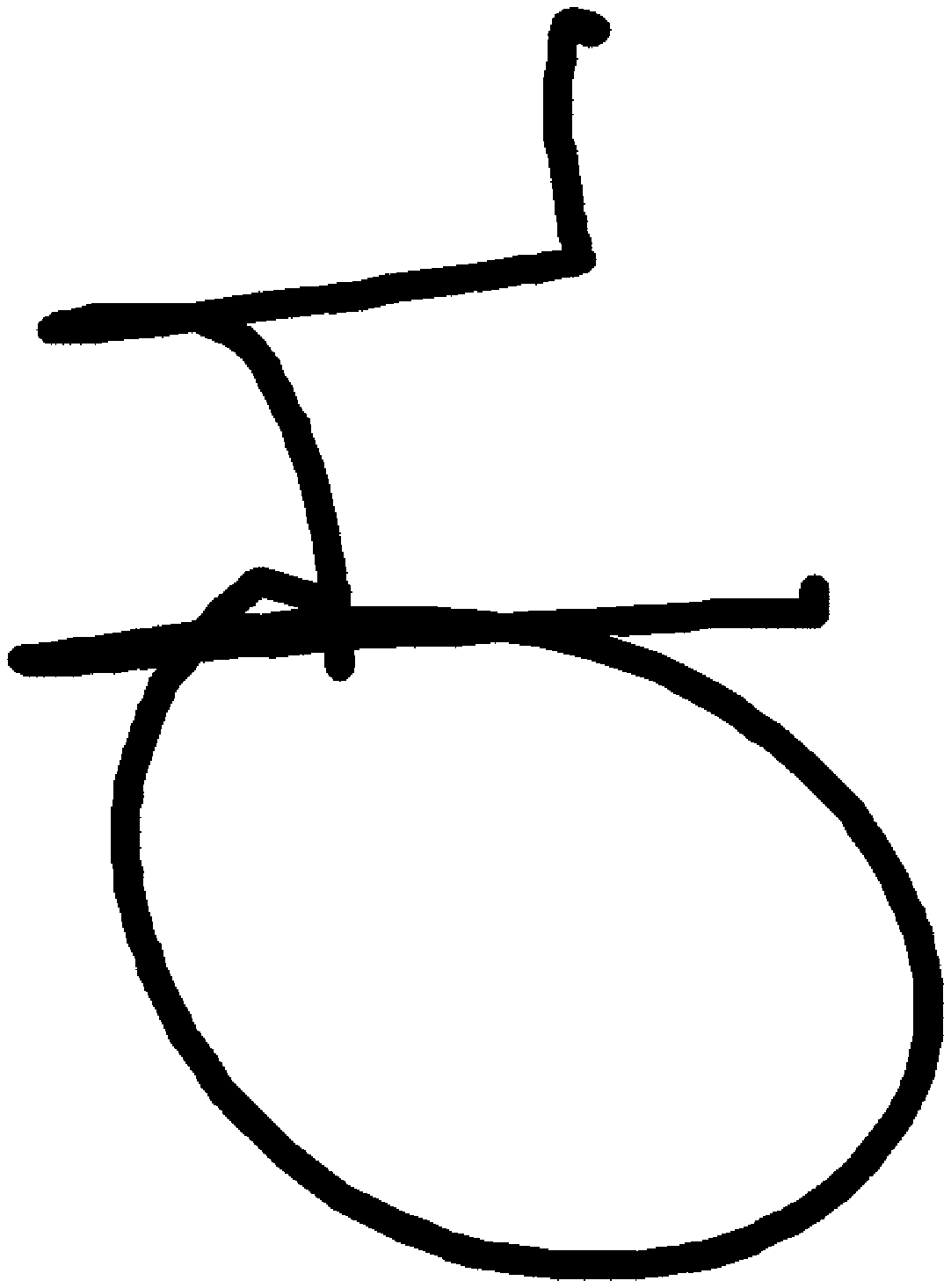} &
			\includegraphics[width=0.5in]{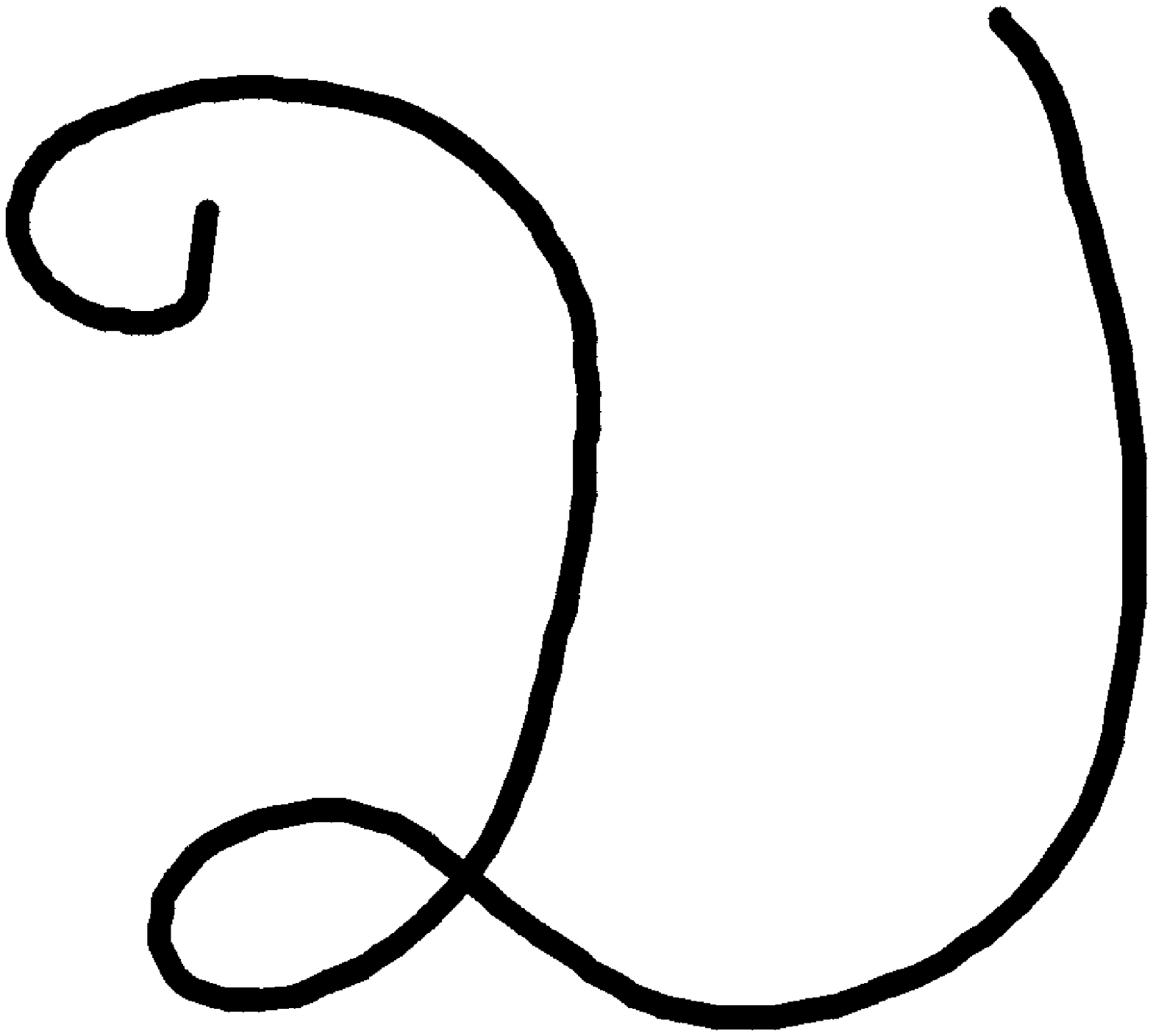} &
			\includegraphics[width=0.5in]{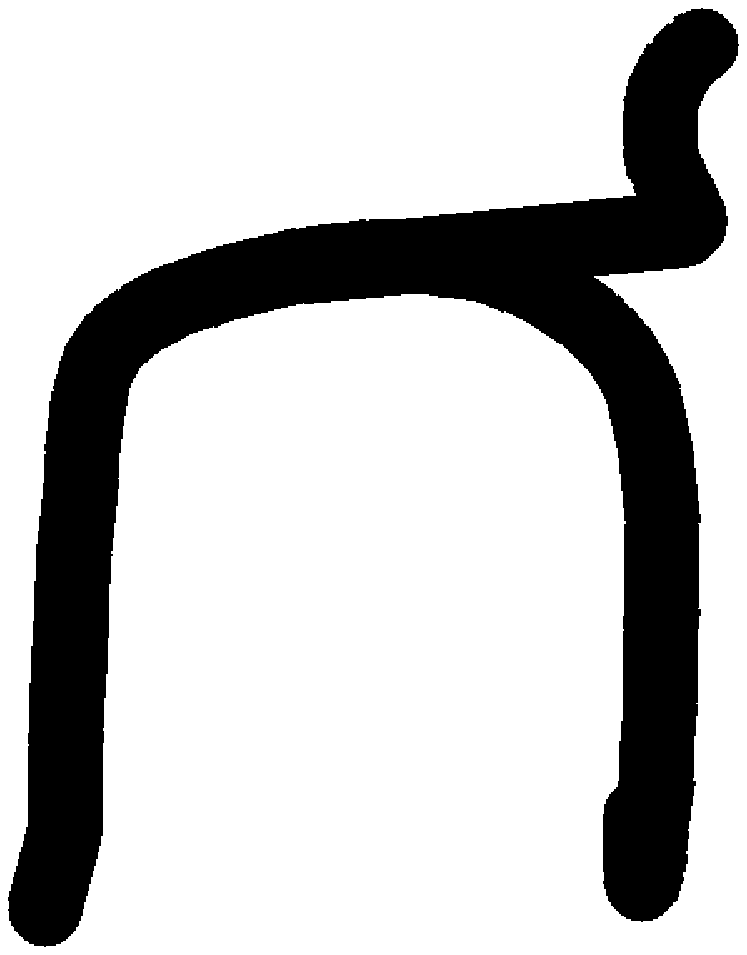} &
			\includegraphics[width=0.5in]{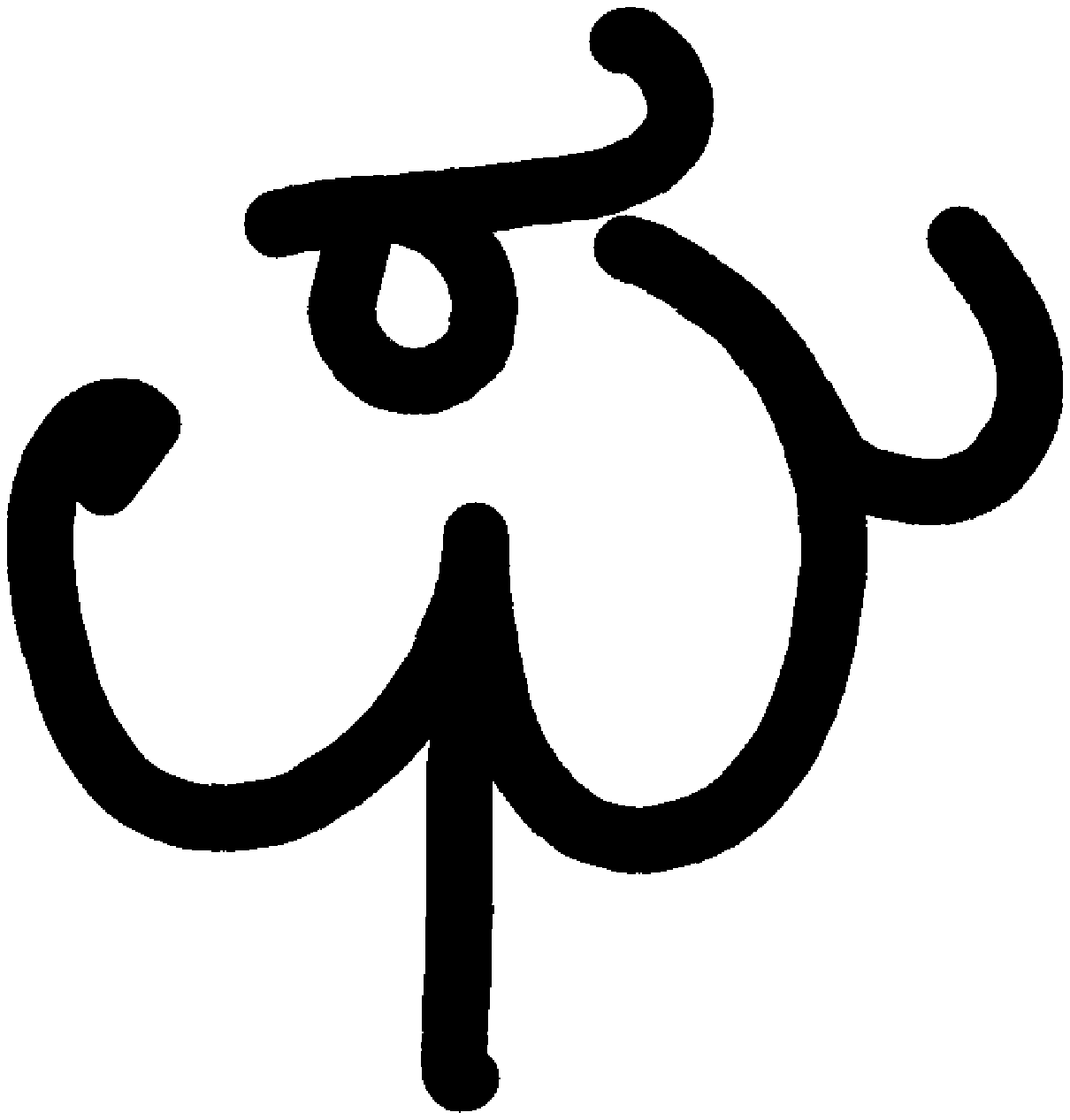} &
			\includegraphics[width=0.5in]{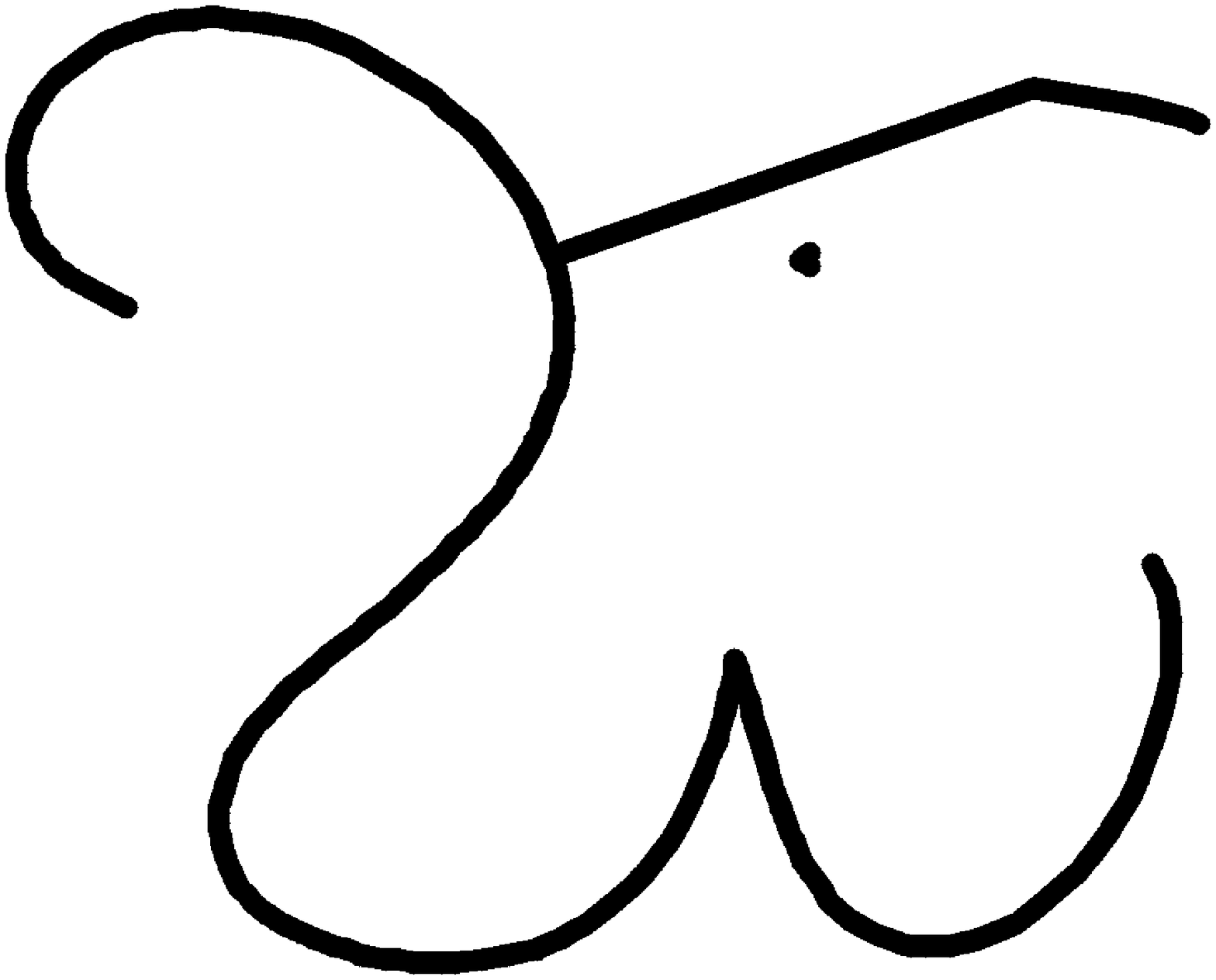} \\
		\end{tabular}
	\end{center}
	\caption{Kannada handwritten characters in the Char74k dataset collected using a tablet PC.} \label{fig:dataset}
\end{figure}

Building an offline HWR system for Kannada involves solving many problems including image preprocessing, character and word segmentation, recognition, integration of language models and context, \textit{etc}. Indic scripts have rich morphological structure where in a character can lead to many other characters after morphological changes and agglutination. This results in a large number of classes. Segmentation, even at the character level, becomes an important task to obtain a high accuracy. In this paper, we propose a method to reduce the number of classes by character segmentation and show that it results in better character recognition. The character recognizer is a building block for freeform handwriting recognition since the same models can be used to recognize words.

Explicit segmentation is generally the first step in handwriting recognition followed by recognition \cite{rahman:impexp, agr:kannada}. In online recognition systems, the coordinates of the points of writing are stored as a function of time, i.e., the order of strokes made by the writer is readily available. This extra information helps in performing explicit segmentation. In printed text, there are no overlapping characters and this makes segmentation relatively easy. However, explicit segmentation of the characters is not popular for offline handwritten data since time-based information is not available and also because of the increased variability as compared to printed text.

Our contribution is to extend the word level implicit segmentation to the character level by taking advantage of the agglutinative nature of Kannada characters. That is, the system searches a character for simpler shapes. We show that this technique improves character recognition accuracy not only by reducing the number of classes, but also by increasing the training samples. The system is based on a sliding window approach: Feature vectors are extracted from fixed width windows which shift column-by-column from left-to-right. The sequence of vectors are modeled with continuous density HMMs. HMM framework allows for simultaneous training and segmentation of these shapes using the features extracted from character images.

The paper is organized as follows: Section~\ref{sec:featextr} describes the feature extraction methodology. These features are used to train the HMM models as described in Section~\ref{sec:HMM}. The dataset and the intuition for segmentation is described in Section~\ref{sec:dataset} followed by experiments and results in Section~\ref{sec:exp}. We also discuss parameter selection and compare with existing systems in this section. Concluding remarks are drawn in Section~\ref{sec:conc}.

\section{Feature extraction} \label{sec:featextr}

The Chars74k dataset \cite{de2009character} contains 657 characters of the Kannada script with 25 samples for each character. These characters are binary images. A sequence of observation vectors are fed as input to the HMMs. These vectors are obtained by feature extraction. We use left-to-right HMMs (discussed in Section~\ref{sec:HMM}). Hence we use a sliding window of fixed width which shifts column by column from left-to-right. At each position, feature vectors are extracted.

\begin{figure}[!t]
	\centering
	\includegraphics[height=1.5in]{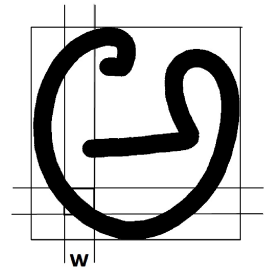}
	\centering
	\caption{Features extracted from cells contained in vertical strips of width $w$. Each vertical strip has $h$ such cells.} \label{fig:featextr}
\end{figure}

The features extracted from this binary image are gradient-based descriptors. Such features have found to be useful in handwritten text recognition \cite{srikant}, human detection \cite{humandet} and hand gesture recognition \cite{handrec}. The idea behind using these features is that local shapes can be characterized using edge directions or by the distribution of local gradient intensities without knowing the precise locations of the corresponding gradient points and edges.

The features are extracted only from the region of the image which contains foreground pixels. In Kannada all the characters are of nearly the same height. Hence, we rescale the isolated character images to a standard height. We implement the algorithm by first dividing this image into vertical strips of width $w$ pixels. We divide each strip into $h$ regions which we call \textit{cells}. In each cell, we compute the histogram of gradient directions over the pixels of the cell. These directions are specified by orientation \textit{bins} evenly spaced over $0^0$ to $360^0$. The combined histogram entries form the feature.

The parameters $w$ and $h$ are fixed by validation. We found that 5 \textit{bins} are sufficient for a binary image. Each pixel in a cell accounts for a weight being added to one of the histogram channels. A fixed weight of one indicating existence in that particular channel is used. We calculate the gradients using a simple $\left[-1~~0~~1~\right]$ mask in both the X and Y directions without any Gaussian smoothing. We found no improvement by using derivative of Gaussian (DoG) kernels to calculate the gradient.

We also build HMM models using hisogram of oriented gradients (HOG) features. The weights used for this feature is the magnitude of the gradient intensity. Comparison of the two different features have been shown in Section~\ref{sec:exp}.

\section{HMM training and recognition} \label{sec:HMM}

\begin{figure}[b]
	\centering
	\includegraphics[height=1in]{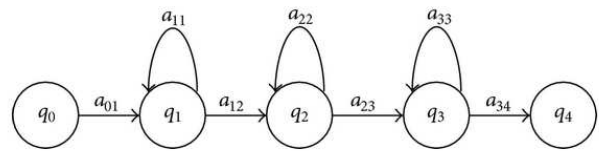}
	\centering
	\caption{An example of a 3 state left-to-right HMM with no state skips} \label{fig:lrhmm}
\end{figure}

The input to our HMM model is a sequence of observation vectors which we obtain from the images of characters (as discussed in Section~\ref{sec:featextr}). We train a different HMM model for each class. The recognition task involves finding a character $\hat{C}$ which maximizes the aposteriori probability of the class ($C$) given the observation sequence ($O$)
\begin{equation} \label{eq:1}
\hat{C} = \operatorname*{arg\,max}_C p(C/O),
\end{equation}
where C could be a vowel, a base class or a modified class as defined in Section~\ref{sec:dataset}. We rewrite Equation~(\ref{eq:1}) using Bayes Theorem as
\begin{equation} \label{eq:2}
\hat{C} = \operatorname*{arg\,max}_C \dfrac{p(O/C)p(C)}{p(O)}.
\end{equation}
$P(O/C)$ is the probability of the observation sequence $O$ being generated by a character $C$. Since $p(O)$ does not depend on class information, Equation~(\ref{eq:2}) is same as
\begin{equation} \label{eq:3}
\hat{C} = \operatorname*{arg\,max}_C p(O/C)p(C).
\end{equation}
Since this is a single character recognition problem, $p(C)$ is assumed to be a uniform distribution over the lexicon (all character classes) and hence Equation~(\ref{eq:3}) is simply
\begin{equation}
\hat{C} = \operatorname*{arg\,max}_C p(O/C).
\end{equation}

This probability $p(O/C)$ is modeled using HMMs. A HMM can be thought of to be a probability density function over a sequence of observations. We use continuous density functions \cite{rabtut}. The emission probabilities of the HMMs are modeled with Gaussian mixtures. In our experiments all models have left-to-right topology i.e only transitions to the next state and self transitions are allowed. A character model is a concatenation of these class models. When the final state of a class is reached, there is either a self transition or a transition to the first state of model of the next class. The number of states of the HMM is represented by $S$ and the number of components in the Gaussian Mixture Model (GMM) by $G$. The assumption made is that each model has the same number of states and same number of components in the GMM.

The models are trained using the Baum-Welch algorithm which is a specific case of the Expectation-Maximization algorithm. It maximizes the likelihood of the training set given the models \cite{baummax}. This algorithm allows for implicit segmentation. We do not segment each class and then train individual models but instead apply this algorithm to the characters. The model training is an iterative process. It first segments all the characters into individual classes using existing model parameters. Based on this segmentation, the parameters are again re-estimated. This process continues till convergence. Thus the boundary of segmentation need not be specified.

The recognition process involves finding the class or the sequence of classes with the highest probability for the observation sequence. This class or sequence of classes is nothing but a character and the list of characters forms the lexicon. The Viterbi algorithm is used to recognize one of the characters in the lexicon \cite{viterbi}. This gives the best likelihood $\lambda$ that can be obtained by following a unique state sequence with the features extracted from the handwritten data. The state sequence giving the highest value of $\lambda$ is selected as the result of the input handwritten character.

\section{Dataset} \label{sec:dataset}

\begin{figure}[!t]
	\centering
	\includegraphics[height=2.15in]{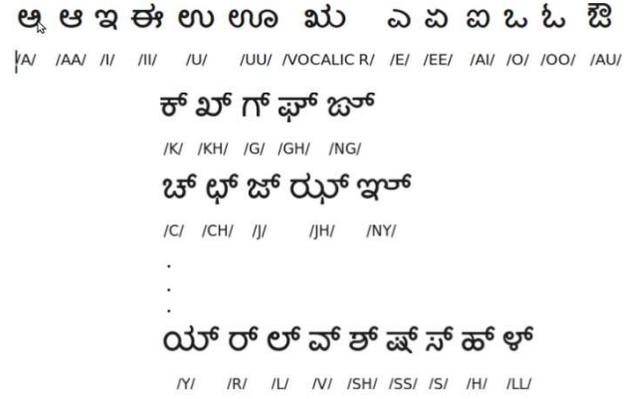}
	\caption{13 vowels and some consonants of the Kannada script} \label{fig:vowcons}
\end{figure}

The Chars74k dataset contains 657 characters of the Kannada script collected using a tablet PC. There are 25 samples for each character and only offline data is available. The stroke thickness of the samples is not uniform, which is the case in handwritten text.

\begin{figure}[b!]
	\centering
	\includegraphics[height=0.33in]{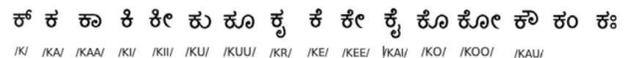}
	\caption{CV combinations for /K/} \label{fig:CVeg}
\end{figure}

\begin{figure}[t!]
	\centering
	\includegraphics[height=0.5in]{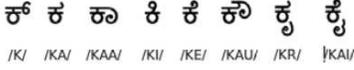}
	\caption{Base classes of /k/} \label{fig:baseclass}
\end{figure}

\begin{figure}[t!]
	\centering
	\includegraphics[height=0.22in]{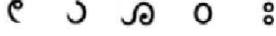}
	\caption{List of modifiers} \label{fig:modifiers}
\end{figure}

\begin{figure}[t!]
	\centering
	\includegraphics[height=0.5in]{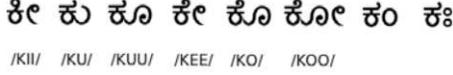}
	\centering
	\caption{The effect of modifiers on some of the base classes of /k/} \label{fig:modclass}
\end{figure}

Modern Kannada script has 13 vowels, 34 consonants and two other letters, namely the anuswara \includegraphics[width=0.12in]{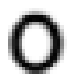} and visarga \includegraphics[width=0.12in]{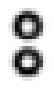}. All consonants combine with all the vowels to form consonant-vowel combinations (CV). In addition, there are 10 numerals. Our work describes initial efforts at developing a recognition system with implicit segmentation of the characters using Hidden Markov Models. We have considered vowels, consonants and consonant-vowel combinations which are available in the dataset. The characters shown in Fig.~\ref{fig:vowcons} to Fig.~\ref{fig:modclass} are not handwritten and shown to describe the intuition behind the segmentation. These are not characters from the dataset.

In Figure~\ref{fig:CVeg}, we have a list of all the CV combinations for the consonant /K/. All consonants combine with the vowels /A/, /AA/, /I/, /E/, /AU/, /VOCALIC R/ and /AI/ (refer Figure~\ref{fig:vowcons}) to form \textit{base classes}. Figure~\ref{fig:baseclass} is a list of the \textit{base classes} of the consonant /K/. These characters can not have further left-to-right segmentation into simpler shapes. Since we use a left-to-right sliding window, it is currently possible to identify only left-to-right segmentation.

In Figure~\ref{fig:modifiers}, we have the list of \textit{modifiers} which join at the right side of some of the base classes to form the list of characters shown in Figure~\ref{fig:modclass}. We refer to these as the \textit{modified classes}. These modifiers are common to all the consonants. The base classes and the modified classes together form all the CV combinations. We therefore have 8 base classes for a consonant and 5 modifiers required to form the modified classes.

As a result of the previous split we have 13 vowels, 34 consonants, 8 base classes, 5 modifiers and 10 numerals. Hence the total number of classes is given by:
\begin{displaymath}
13 + (34 * 8) + 5 + 10 = 300
\end{displaymath}

The problem is thus reduced to a 300 class classification task. 

The Chars74k dataset contains 657 characters with 25 samples for each character. Some of the characters are obsolete in the present Kannada script (\includegraphics[width=0.15in]{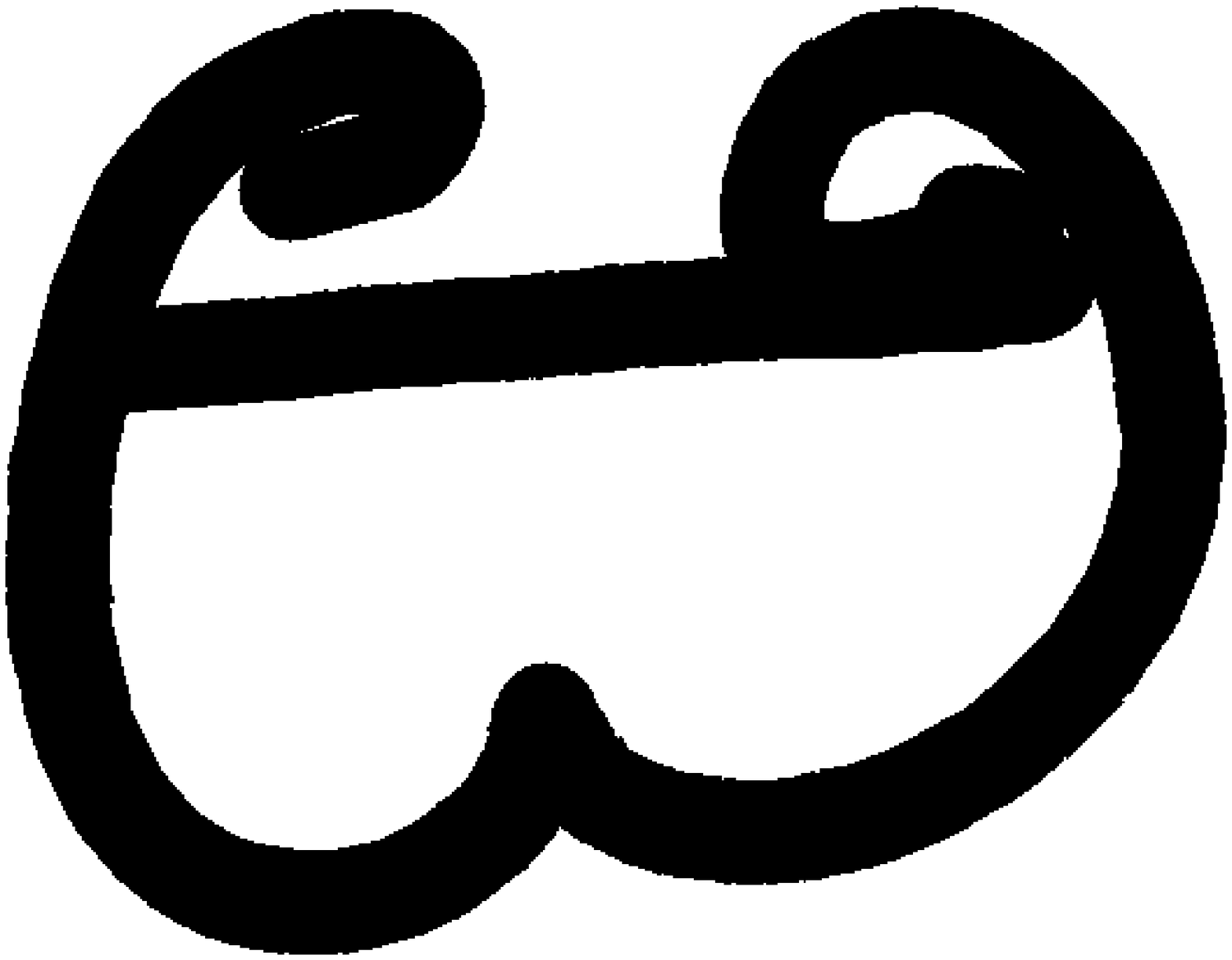},\includegraphics[width=0.19in]{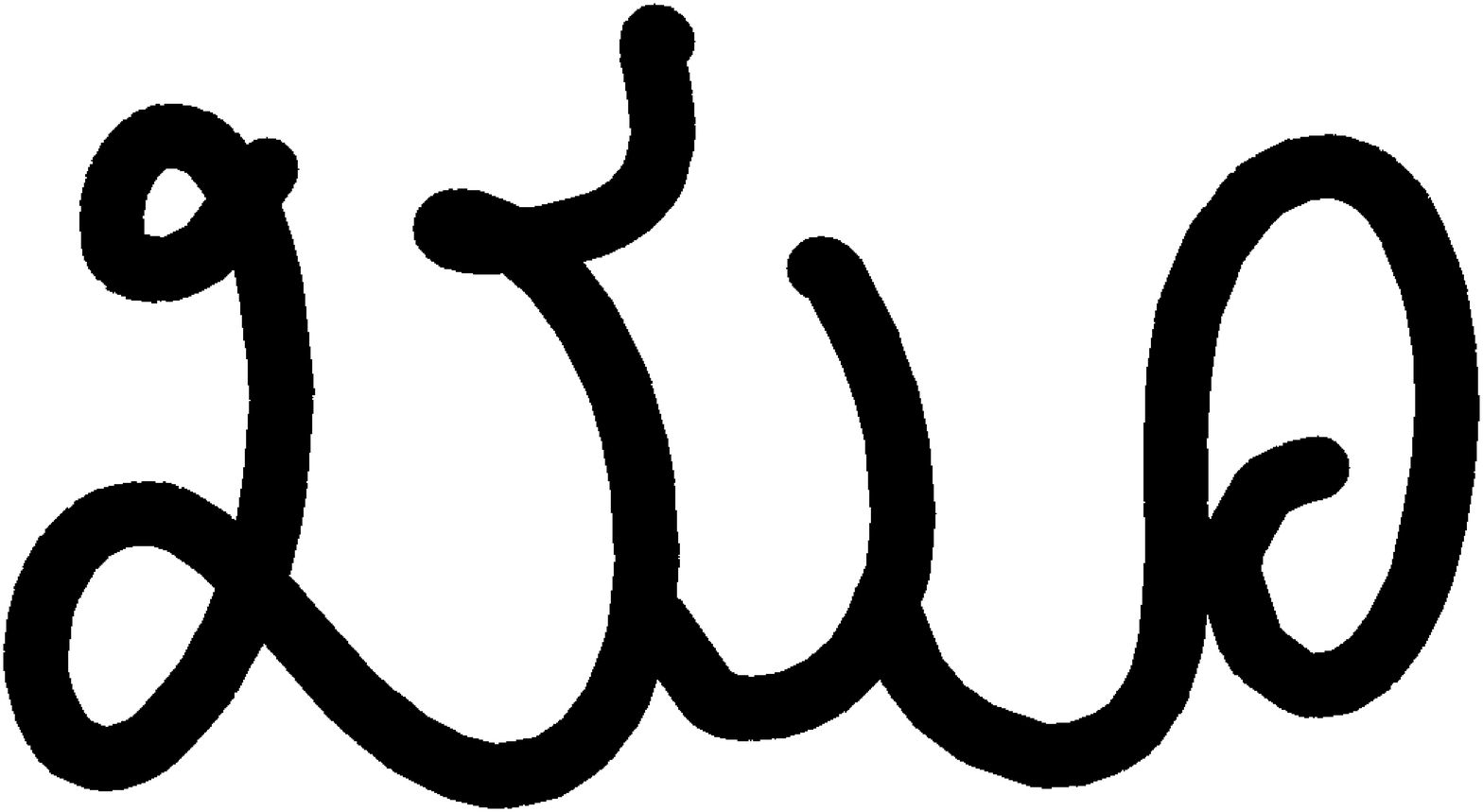},\includegraphics[width=0.15in]{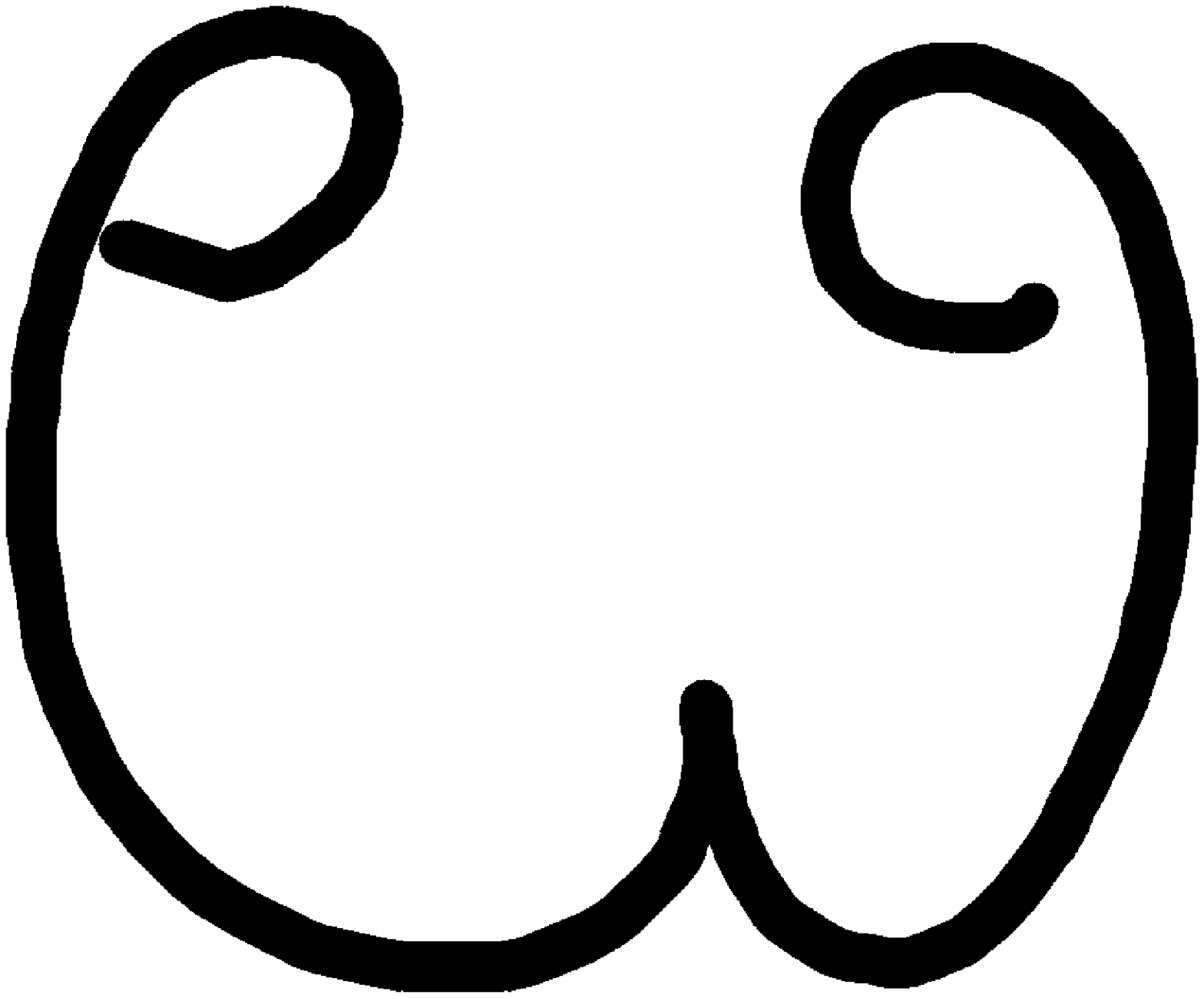}). Hence we consider only 569 characters out of the 657. These are either vowels, base classes, modified classes or numerals. We use the segmentation methodology described and build only a 300 class classifier to classify these 569 characters. 

We not only reduce the number of classes by this method but we also increase the number of training samples for most classes since base classes are repeated in the modified classes of the corresponding consonant and modifiers are common to all CV combinations of different consonants.

\section{Experiments and results} \label{sec:exp}

The Chars74k dataset contains 25 samples of each character. For all our experiments we divide the dataset into 15 samples for training, 5 for validation and 5 for testing. For the 15 - 5 training - validation split, we have done a 4-fold cross validation. We extract HOG features and gradient-based features from the image representation and we use the validation data to select feature parameters $w$ and $h$ (See Section~\ref{sec:featextr}) and HMM parameters such as number of states ($S$) and number of Gaussians ($G$). We use the Hidden Markov Model Toolkit (HTK) for HMM parameter estimation and evaluation \cite{Young2006}.

\subsection{569 classes}

We consider each of the 569 characters as separate classes to form a benchmark for later comparison. We model each class separately using a left-to-right HMM (569 HMMs). We considered HOG and gradient-based features and tuned all the parameters in the validation set. The best classification accuracy obtained was 50.61\%. This result corresponds to HMM parameters $S = 15$ and $G = 4$ and gradient-based features with $h = 8$ and $w = 8$.

The total number of parameters in this system is too high. The system has 569 HMMs each with 15 states; each state modeled with a 4 component GMM. The training accuracy is 98.21\% which shows that the system overfits the data. This is expected because of the large number of parameters and the low number of training samples.

\subsection{Implicit segmentation}

The 569 characters is split into 300 classes as described in Section~\ref{sec:dataset}. Baum-Welch algorithm identifies the segmentation and trains the parameters. The Viterbi algorithm finds the best sequence of classes during testing. In our experiments we start with a simple set of single component GMMs and then iteratively refine them by using multiple mixture component Gaussian distributions. The HMM parameters are re-estimated in these iterations. For a single Gaussian component we re-estimate the parameters twice and then double $G$ and continue the procedure.

\subsubsection{Parameter selection}

\begin{figure}[b!]
	\centering
	\includegraphics[height=2.2in]{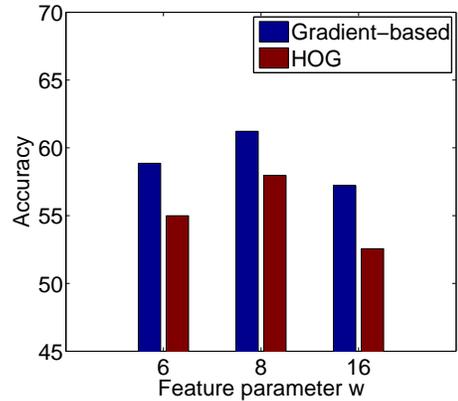}
	\centering
	\caption{Validation accuracy for the two features and different feature parameters. Left bar represents gradient-based features and right bar represents HOG features. (color online)} \label{fig:feat}
\end{figure}

\begin{figure}[!t]
	\centering
	\includegraphics[height=2.2in]{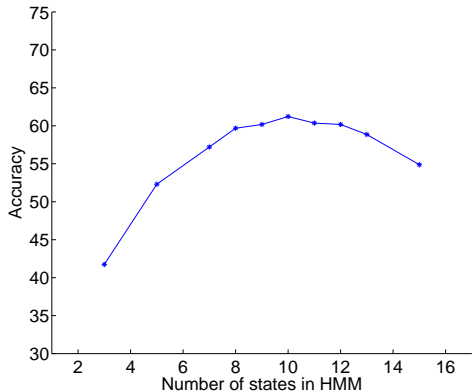}
	\centering
	\caption{Validation accuracy plotted against number of states of HMM. The accuracy decreases after a stage due to overfitting.} \label{fig:numstates}
\end{figure}

It is important to avoid overfitting of the training data. We select our parameters for the features, number of states in the HMM, and number of components in the GMM optimally in the following way.

We try different parameters $w$ and $h$ for feature extraction for the gradient-based and HOG features. Validation is performed for feature parameters $w$ and $h$ and HMM parameters $S$ and $G$.% and the accuracy is shown in Table~\ref{tab:tabfeat}.

Based on validation, the parameters chosen are
\begin{displaymath}
S = 10; G = 4; w = 8; h = 8;
\end{displaymath}

The bar graph in Figure~\ref{fig:feat} shows the comparison of validation accuracy for the two features for varying $w$. We found best results for $h$ equal to $8$. Hence we choose gradient features for all the subsequent experiments. 

The graph in Figure~\ref{fig:numstates} shows the best classification accuracy on the validation data with respect to the number of states $S$ for the optimal values of $w$ and $h$. The reduction in validation accuracy for $S$ greater than $10$ is a consequence of overfitting.

\begin{figure}[t]
	\centering
	\includegraphics[height=2.2in]{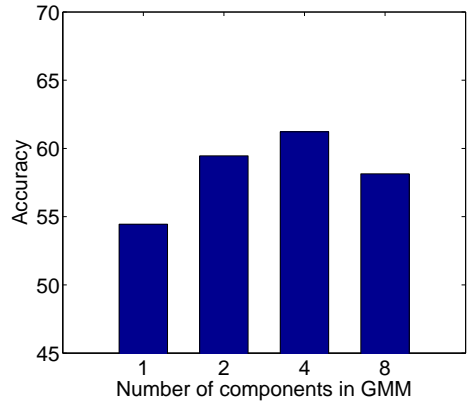}
	\centering
	\caption{Validation accuracy plotted against number of components in the GMM. The accuracy decreases after a stage due to overfitting.} \label{fig:numgauss}
\end{figure}

The graph in Figure~\ref{fig:numgauss} shows the performance of the classifier with increasing $G$ for the validation data. A single Gaussian is not capable of modeling the data well enough but using 8 components in the GMM is leading to overfitting.

The best validation accuracy of 61.22\% was obtained for number of states $S = 10$. Using implicit segmentation at the character level results in reduced number of classes to train and also more number of samples for some classes. This reduction in the number of class has not only increased the accuracy but also for a simpler model. The new model has only 10 states as compared to 15 for the classifier used in the previous experiment. The new model has fewer parameters which results in faster training and recognition time.

The accuracy on the test data we obtained with these parameters is 61.0\%. This is around 10\% higher than treating the 569 characters as independent classes. This is the result of HMMs exploiting the underlying structure and increased training samples as explained above.

Furthermore, we compare our system with a nearest neighbour classifier based on DCT features (DCT - NN) \cite{kumar:scene}. We also performed experiments to directly compare with this system which considers all the 657 characters in the dataset. The 25 samples are split into 12 training samples and 13 test samples. The results are listed in Table~\ref{tab:finalcomp}.

\begin{table}[h]
	\caption{Comparisons with the existing system} \label{tab:finalcomp}
	\begin{center}
		\begin{tabular}{ | c | p{2.2cm} | p{3cm} |}
			\hline
			DCT - NN &  HMMs without segmentation  &  HMMs with implicit segmentation \\ \hline\hline
			33.3\% & \multicolumn{1}{c|}{39.79\%} &  \multicolumn{1}{c|}{\textbf{49.22\%}} \\ \hline
		\end{tabular}
	\end{center}
\end{table}

Typically, state-of-the-art handwriting recognition systems for other languages have better performance \cite{hmm:eng}. We further analyze the system using learning curves. Learning curve is the plot of training and/or validation accuracy as the number of training samples increases. In this experiment, we increase the number of training samples from 5 till 15 and find the corresponding validation and training set accuracy. When the number of training samples is low training accuracy will be high (overfitting) and validation accuracy is low (poor generalization). As the number of training samples increase, the training accuracy will decrease and validation accuracy will increase. A good machine learning system will have comparable training and validation accuracy.

\begin{figure}[t]
	\centering
	\includegraphics[height=2.5in]{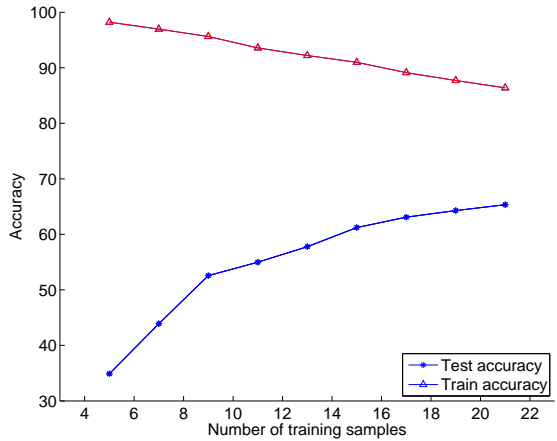}
	\centering
	\caption{Improvement in test accuracy with increase in training samples indicates the requirement for more training data.} \label{fig:trainsamp}
\end{figure}

Figure~\ref{fig:trainsamp} shows the learning curve for our system. It can be seen that the training accuracy is decreasing and the validation accuracy is increasing. However the trend shows that more training samples could be added to increase the accuracy. 

Accordingly, we collected 6 more samples for each of the 500 characters and added it to the training data\footnote{The additional data collected is available at \url{https://sites.google.com/site/manasijvenkatesh/research-projects/hwr}}. The accuracy we obtained with the increased data is 65.34\%. An improvement of 4\% is observed with the addition of just 6 samples which further reiterates our analysis that the poor accuracy is due to the lack of data. The graph in Figure~\ref{fig:trainsamp} includes results with the additional data.

\section{Conclusions} \label{sec:conc}

Our work presents a recognizer system for the offline recognition of Kannada handwritten characters based on continuous density Hidden Markov Models. The Kannada script has a large number of characters most of which are morphological changes of a base character. Hence, segmentation becomes an important task in reducing the complexity of the classifier. Since we use HMMs for classification, explicit segmentation is avoided and is a byproduct of the recognition. This implicit segmentation technique at the character level has showed improved results by reducing the number of classes along with the increase in the training samples. In addition, it also reduces difficulty in the task of data collection as the data need not include segmentation boundaries.

Very few works have been dedicated to the offline recognition of Kannada handwritten text. This is, to the authors' best knowledge, the best reported accuracy on the Chars74k dataset. It suggests that the features are robust to high variability in input pattern. However, this accuracy is inferior compared to state-of-the-art handwriting recognition systems of other scripts. The reason, as evident from the learning curves, is the lack of training data. However, adding data improved the accuracy. An important future work is to collect more data. In addition, this data collection effort should include Consonant Consonant Vowel (CCV) combinations of characters which are currently not present in the dataset.

We currently exploit the agglutinative nature of the script in the horizontal direction, but not in the vertical direction. Potentially, one can look into techniques such as HMM adaptation to address the issue. Future work also involves including CCV combinations to build a complete system and to further improve the accuracy by introducing a language model (lexicon containing a large number of words). This technique can be employed to the recognition of other Indic scripts which have agglutinative nature (Devanagari, Telugu).

\bibliographystyle{IEEEtran}%{abbrv}
\bibliography{bibhwr}  

\end{document}